\documentclass[letterpaper, 10 pt, conference]{ieeeconf}  

\IEEEoverridecommandlockouts                              

\usepackage{graphics}
\usepackage{epsfig}
\usepackage{times}
\usepackage{amsmath}
\usepackage{amssymb}
\usepackage{amsfonts}
\usepackage{cite}
\usepackage[ruled,linesnumbered]{algorithm2e}
\usepackage{graphicx}
\usepackage{textcomp}
\usepackage{pdfpages}
\usepackage{caption}
\usepackage{float}

\makeatletter
\let\NAT@parse\undefined
\makeatother

\usepackage[table,xcdraw]{xcolor}

\makeatletter
\let\NAT@parse\undefined
\makeatother

\definecolor{CiteBlue}{RGB}{30, 90, 150}

\usepackage[
    colorlinks=true,
    linkcolor=red,        
    citecolor=CiteBlue,   
    urlcolor=magenta      
]{hyperref}

\usepackage{booktabs}
\usepackage{multirow}
\usepackage[table,xcdraw]{xcolor}
\usepackage[normalem]{ulem}
\useunder{\uline}{\ul}{}

\title{\LARGE \bf
HiSfM: Disambiguating Structure-from-Motion via Scaffold-Anchored Hierarchical Reconstruction
}

\author{
Ziding Zhao$^{1,2}$ \quad
Hainan Cui$^{1,2\dagger}$ \quad
Peilin Tao$^{1,2}$ \quad
Shuhan Shen$^{1,2}$\\
$^{1}$Institute of Automation, Chinese Academy of Sciences\\
$^{2}$School of Artificial Intelligence, University of Chinese Academy of Sciences
}

\makeatletter
\renewcommand{\footnoterule}{%
  \kern -3pt
  \hrule width 0.4\columnwidth height 0.4pt
  \kern 2.6pt
}
\makeatother

\begin{document}

\setlength{\textfloatsep}{8pt plus 1pt minus 2pt}
\setlength{\dbltextfloatsep}{8pt plus 1pt minus 2pt}
\setlength{\floatsep}{6pt plus 1pt minus 1pt}
\setlength{\dblfloatsep}{6pt plus 1pt minus 1pt}
\setlength{\intextsep}{8pt plus 1pt minus 2pt}
\setlength{\abovecaptionskip}{3pt}
\setlength{\belowcaptionskip}{0pt}

\twocolumn[{%
    \renewcommand\twocolumn[1][]{#1}%
    \maketitle
    \vspace{-1.6em}  

    {%
    \centering
    \captionsetup{type=figure}
    \includegraphics[page=1,width=\textwidth,height=7cm,keepaspectratio]{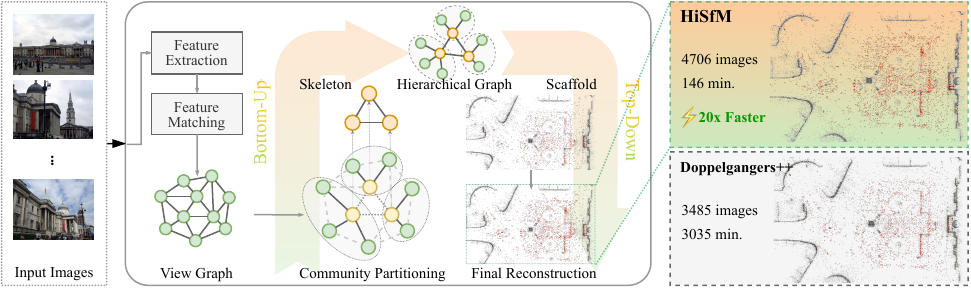}

    \captionof{figure}{\textbf{HiSfM Overview}: We propose a hierarchical scaffold-anchored structure-from-motion pipeline, which builds a skeleton from \textit{bottom-up}, followed by \textit{top-down} scaffold reconstruction and iterative registration and triangulation to achieve complete results. The results on example large-scale scene \textit{Trafalgar}\cite{wilson_robust_2014} shows that our method reconstruct the scene with better completeness while runs 17 times faster than previous state-of-the-art method Doppelgangers++ \cite{xiangli_doppelgangers_2025}.}
    \label{fig:intro}
    \par
    \vspace{0.6em}
    }%
}]

\begingroup
\renewcommand{\thefootnote}{\fnsymbol{footnote}}
\footnotetext[2]{Corresponding author}
\endgroup


\begin{abstract}

    Structure-from-Motion (SfM) is a fundamental tool for sparse 3D reconstruction with broad impact in robotics and vision, supporting mapping, localization, and large-scale scene modeling. However, conventional pipelines often fail under hard visual ambiguity caused by repeated or symmetric structures, and incur heavy computational cost due to redundant cameras and constraints. We present HiSfM, a hierarchical coarse-to-fine SfM framework that improves robustness and efficiency through scaffold construction. HiSfM first forms strong local communities using geometrical induced heuristics, then connects communities with a compact yet strong skeleton by packing edge-disjoint spanning trees (EDST) while verifying skeletal edges with a two-view disambiguator. We reconstruct a stable scaffold on this verified skeleton, serving as an anchor to capture the essence of the scene, and subsequently absorb remaining images via efficient registration and triangulation for further refinements. Experiments on ambiguity-focused benchmarks and general datasets show that HiSfM prevents ambiguity-induced failures while substantially reducing runtime compared to previous methods, and improves completeness over aggressive sparsification methods. Code is available at \href{https://github.com/3dv-casia/HiSfM}{https://github.com/3dv-casia/HiSfM}.

\end{abstract}

\section{Introduction}
    \label{intro}

    Structure-from-Motion (SfM) estimates camera poses and sparse 3D points from a collection of images, and is a key building block not only for large-scale content creation but also for robotics and autonomous systems, enabling map building, visual SLAM, and 6-DoF localization~\cite{klein_ptam_2007,campos2021orb,sattler_prioritized_2016,sattler_benchmarking_2018,agarwal_building_2009,snavely_modeling_2008,mildenhall_nerf_2020,kerbl_3dgs_2023}. Modern systems such as COLMAP~\cite{schonberger_structure--motion_2016} and GLOMAP~\cite{pan_global_2024} have made SfM broadly usable across diverse capture conditions and datasets, and SfM maps built with COLMAP are widely used as reference reconstructions in visual localization and robotics-oriented benchmarks~\cite{sattler_benchmarking_2018}.
    
\begin{figure}
        \centering
        \includegraphics[width=1\linewidth]{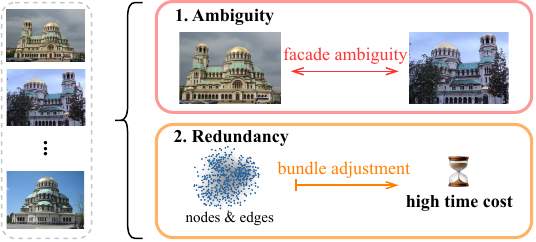}
        \caption{
        Two major challenges in SfM: 1) Visual ambiguity from repetitive patterns; 2) Nodes and edges redundancy in the view graph.
        }
        \label{fig:two_challenges}
    \end{figure}

    Despite this progress, SfM still faces two fundamental challenges—robustness under hard \emph{visual ambiguity} and efficiency under \emph{redundancy}, as shown in Fig.~\ref{fig:two_challenges}. First, SfM can fail catastrophically under hard visual ambiguity. In scenes with repeated or symmetric structures, visually plausible correspondences may remain consistent with two-view epipolar geometry and survive standard RANSAC-based verification~\cite{fischler_random_1981}. Once such ambiguous relations enter the view graph~\cite{arrigoni2024direct}, they can mislead pose estimation and triangulation, and the resulting errors may propagates and subsequent bundle adjustment, producing hallucinated reconstructions or duplicate structures. Second, SfM pipelines incur heavy computational overhead as reconstructions grow. In particular, bundle adjustment (BA) is the main bottleneck and is often described as scaling cubically, $\mathcal{O}(n^3)$, with the number of cameras in dense settings~\cite{triggs1999bundle}. This cost is exacerbated by redundancy in the view graph from two sources: 1) Node redundancy, where dense captures may contain many near-duplicate images that contribute little additional geometric information beyond a representative subset; 2) Edge redundancy, where an overly dense set of verified pairs introduces excessive constraints, increasing problem size and coupling and slowing BA. While bundle adjustment is often analyzed through the size of the reduced camera system, the practical runtime of incremental SfM is also affected by the number of observations, graph sparsity, and repeated local/global refinement during reconstruction. Therefore, our goal is not simply to remove edges, but to build a compact and reliable scaffold that preserves global connectivity while avoiding unnecessary verification and optimization on redundant cameras and constraints.

    A variety of approaches address these challenges, but they often fall on opposite ends of a robustness--efficiency trade-off. On the robustness side, learning-based visual disambiguation methods such as Doppelgangers~\cite{cai_doppelgangers_2023} and Doppelgangers++~\cite{xiangli_doppelgangers_2025} explicitly predict whether a visually similar pair corresponds to the same 3D surface. While highly effective in suppressing ambiguity-induced failures, their cost becomes prohibitive when applied broadly across many candidate edges in large view graphs because they brutally traverse every image pair. On the efficiency side, graph-selection and sparsification strategies accelerate SfM by reconstructing only a compact subset or by aggressively reducing edges---including skeletal subset selection~\cite{snavely_skeletal_2008}, randomized reconstruction schedules~\cite{havlena_randomized_2009}, efficient pose-graph initialization~\cite{barath_efficient_2021}, and triplet-driven graph sparsification such as CamTrip~\cite{manam_leveraging_2024}. These methods can significantly reduce runtime cost, but they may oversimplify global connectivity. When the retained graph becomes too sparse, it can oversplit the scene or remove the few critical links needed to propagate registration across ambiguous structures, thus reducing completeness and reliability.
    
    In this work, we argue that the key is \emph{where} to spend the robustness budget. Visual ambiguity is most harmful on global propagation paths that connect otherwise well-supported local groups of images. A single incorrect inter-group bridge can corrupt large parts of the reconstruction, whereas dense intra-group relations are often mutually reinforced by many consistent views. This motivates a hierarchical design that separates local community formation from globally critical connectivity, and applies expensive disambiguation selectively to the latter.     

    \textbf{Contributions}: We propose HiSfM, a hierarchical coarse-to-fine SfM framework that improves robustness and efficiency. 
    As shown in Fig.~\ref{fig:intro}, HiSfM first forms strong local communities via lightweight inlier-driven top-$k$ filtering, then builds a compact yet redundant \emph{skeleton} by packing edge-disjoint spanning trees at the community level and selectively verifying only inter-community bridges with a strong two-view disambiguator. We reconstruct a stable scaffold on this verified skeleton and then efficiently attach the remaining images via registration and triangulation. By restricting expensive verification to globally critical bridges and reducing redundancy in the optimized graph, HiSfM mitigates ambiguity-induced drift and accelerates reconstruction. Experiments on ambiguous and general datasets show improved robustness over standard pipelines and competitive runtime with sparsification-based methods while achieving higher completeness.

\section{Related Work}
    \label{relate}
    \begin{figure*}[htbp]
        \centering
        \includegraphics[width=1.0\textwidth]{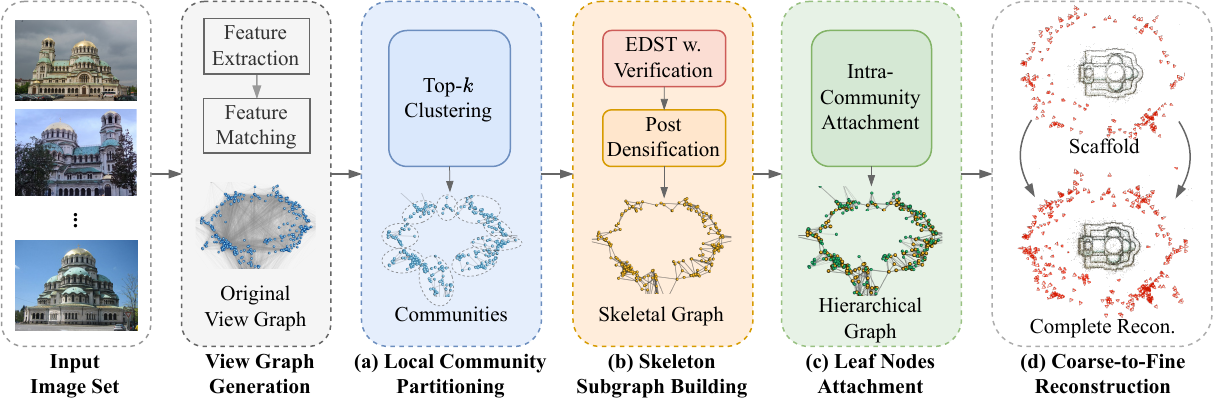}
        \caption{The pipeline of our method: (a) \textcolor{blue}{Blue} nodes indicate view graph nodes generated using top-$k$ clustering (Sec. \ref{sec:local}); (b) \textcolor{orange}{Orange} nodes represent the skeleton, which is constructed by the verified edge-disjoint spanning trees (EDST) generation module (Sec. \ref{sec:skeletal}); (c) Leaf nodes are colored by \textcolor{green}{green} and attaching process is shown in Sec. \ref{leaf};  (d) The structural scaffold is build first, after which a registration–triangulation loop is performed to obtain the final reconstruction (Sec. \ref{sec:recon}).} 
        \label{method_pipeline}
    \end{figure*}
    
    We group prior work along two complementary directions: methods that improve robustness under hard ambiguity by disambiguating incorrect visual relations, and methods that improve efficiency and scalability by reducing the effective problem size or accelerating reconstruction. HiSfM explicitly combines both by allocating expensive disambiguation only to globally critical connections while retaining a compact coarse-to-fine reconstruction schedule.
    
    \textbf{Disambiguating SfM.}
        Early methods addressed ambiguity by reasoning over multiple hypotheses or suppressing unreliable associations using visibility or track context, e.g., sampling and EM-style reasoning for duplicate structures~\cite{roberts_structure_2011} and network/track-based principles for disambiguation~\cite{wilson_network_2013}. Higher-order consistency has also been widely explored: loop or cycle constraints can enforce consistency over the relation graph~\cite{zach_disambiguating_2010}, and geodesic/topological context can help distinguish visually similar views~\cite{yan_distinguishing_2017}. Other methods improve robustness by post-hoc correction of duplicate structures~\cite{heinly_correcting_2014}. More recently, learning-based disambiguators predict whether visually similar pairs correspond to the same 3D surface~\cite{cai_doppelgangers_2023,xiangli_doppelgangers_2025}, while view-graph-level methods select or generate more reliable graphs using optimization, probabilistic modeling, online learning, or cluster-level pose consistency~\cite{damblon2025learning,zhou_pgvs_2026,gan_lvg_2024,gong_cluster_2024}.
        
        Despite their effectiveness, these methods often incur substantial cost when applied broadly. Triplet, cycle, or global view-graph selection methods may need to evaluate many higher-order structures or candidate edges, post-hoc correction requires an already reconstructed model, and pairwise learned disambiguators become expensive when invoked on a large fraction of the view graph. Moreover, methods that globally filter or construct the view graph do not explicitly distinguish locally redundant support from globally critical propagation paths. HiSfM is motivated by this distinction: ambiguous errors are most damaging on inter-submap bridges, whereas intra-submap relations are usually supported by many nearby views. Therefore, instead of globally disambiguating all relations, HiSfM first forms locally supported submaps, verifies only the bridge edges selected for a compact skeleton, and reconstructs this scaffold before attaching the remaining images. This bridge-level verification and scaffold-to-leaf schedule provide a targeted robustness--efficiency trade-off for hierarchical SfM.

    \textbf{Efficient and scalable SfM.}
    To reduce runtime and memory, many systems reduce the effective problem size or restructure the reconstruction schedule. Skeletal-graph approaches select a compact representative subset, reconstruct it first, and then absorb remaining images, yielding large speedups on highly redundant collections~\cite{snavely_skeletal_2008}. Related key-subset and coarse-to-fine strategies similarly prioritize a stable core before completing the reconstruction~\cite{gong_view-graph_2023}. Other work accelerates large-scale SfM through randomized seeding and local-to-global merging, e.g., triplet-based seeding in randomized SfM~\cite{havlena_randomized_2009}, faster pose-graph initialization by reusing information along existing graph paths~\cite{barath_efficient_2021}, and incremental view-graph construction that propagates reliability from strong edges~\cite{cui_view-graph_2021}. View-graph sparsification methods reduce redundancy while aiming to preserve reconstructability; in particular, triplet-based scoring has been used to sparsify graphs while suppressing false edges from repeated structures~\cite{manam_leveraging_2024}. Beyond graph selection, divide-and-conquer and cluster-based pipelines parallelize local reconstructions and merging for scalability~\cite{chen_graph-based_2020, zhu2018very}, covisibility-based scheduling unifies acceleration strategies across sequential and unordered collections~\cite{ye_ec-sfm_2024}, and distributed systems parallelize computation across machines to scale to very large scenes~\cite{baid_distributed_2023}.

    Nevertheless, pure efficiency-driven sparsification can remove globally important connections, leading to oversplitting or reduced completeness when the remaining edges are insufficiently informative or when ambiguity contaminates the retained constraints~\cite{manam_leveraging_2024}. Skeleton/key-subset strategies reduce variables but do not, by themselves, guarantee that the retained subset avoids ambiguity-induced drift, and aggressive reduction may discard the very context needed to disambiguate repeated structures~\cite{snavely_skeletal_2008,gong_view-graph_2023}. Cluster-based and distributed pipelines improve throughput, but their reliability often hinges on robust merging across submodels and sufficient inter-cluster connectivity; failures at the interfaces can still degrade global consistency~\cite{chen_graph-based_2020, baid_distributed_2023}. These limitations motivate combining selective robustness with structured efficiency rather than optimizing only one axis.

\section{Methodology}
    We propose \textbf{HiSfM}, a hierarchical coarse-to-fine SfM framework designed to address the two challenges highlighted in Fig.~\ref{fig:two_challenges}: (i) \emph{visual ambiguity} that can propagate globally through incorrect long-range connections, and (ii) \emph{runtime inefficiency} caused by redundant cameras and constraints that slow bundle adjustment. The central idea is to separate local and global structure in the view graph: we keep dense, lightweight connectivity within reliable local communities, but selectively apply verification only to inter-community \emph{bridges} that control global propagation. We then reconstruct in a coarse-to-fine schedule—first building a stable scaffold on a compact verified skeleton, and then efficiently attaching the remaining images—so that expensive optimization is dominated by a much smaller core before expanding to the full set. We take as input a collection of images  together with the view graph produced by standard feature extraction and matching, as illustrated in Fig. ~\ref{method_pipeline}.

    \subsection{Local Community Partitioning}
        \label{sec:local}
        We begin by partitioning the input view graph into local communities.. Let the input view graph be $G=(V,E)$, where $V=\{I_i\}_{i=1}^n$ denotes the images and each edge $e_{ij}\in E$ is weighted by $w_{ij}$ (the number of geometrically verified inliers). Our goal is to form local groups that are strongly supported by high-confidence relations, so that most redundancy is kept locally where it improves stability, while leaving global connectivity to be handled explicitly in the next stage.
        
        \begin{figure}
            \centering
            \includegraphics[width=0.95\linewidth]{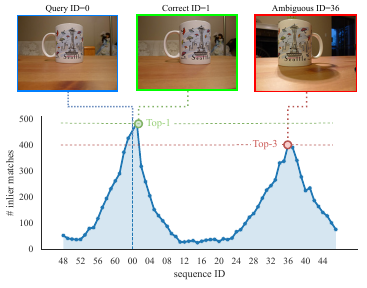}
            \caption{Visualization for number of matches using first image as query in dataset \textit{cup} \cite{yan_distinguishing_2017}. We mark the top-1 neighbour (image ID 1) and the top-3 neighbour (image ID 36).}
            \label{fig:num_inliers}
        \end{figure}
        
        While many community detection and graph clustering methods exist~\cite{blondel_fast_2008,traag_louvain_2019,shi_normalized_2000,rosvall_maps_2008}, we adopt a simple and controllable \emph{top-$k$} strategy tailored to SfM: for each image, we keep only its $k$ strongest neighbors by $w_{ij}$. This produces a sparse subgraph $G_{\mathcal{C}}=(V,E_{\mathcal{C}})$ that preserves strong local support while discarding many weak or redundant relations. We then compute connected components on $G_{\mathcal{C}}$ to obtain the community partition $\mathcal{P}=\{\mathcal{C}_l\}_{l=1}^{m}$, where each $\mathcal{C}_l\subseteq V$ is an community and $m=c(G_{\mathcal{C}})$ is the number of components.

        This simple heuristic is effective because within an community, edges typically have substantially higher inlier support, making them less likely to be spurious and more useful for stable registration, as illustrated in Fig.~\ref{fig:num_inliers}. In practice, top-$1$ filtering yields two benefits: (1) it produces communities with strong mutual support with higher resgistration success rate; (2) it limits unnecessary edge redundancy early, reducing the size of the graph that will drive back-end optimization. Hence, we set $k=1$ for our method.
    
    \subsection{Skeletal Subgraph Building}
        \label{sec:skeletal}
        Given the community partition $\mathcal{P}$, our next step is to derive a compact but robust \emph{skeletal} subgraph that connects communities into a globally traversable structure. This step is where we explicitly guard against ambiguity: incorrect \emph{inter-community} connections can propagate errors across large portions of the reconstruction, so we concentrate expensive disambiguation on these globally critical links.
        
        We define \emph{bridges} as edges that connect nodes from different communities:
        $\mathcal{B} = \left\{ \{i,j\} \in E \;\middle|\; I_i \in \mathcal{C}_a,\ I_j \in \mathcal{C}_b,\ a \neq b \right\}.$
        We then construct the community-level (quotient) multigraph $\bar{G}=(\mathcal{P},\mathcal{B})$, whose vertices are communities and whose edges are bridges with weights inherited from $w_{ij}$. To ensure both compactness and robustness, we seek controlled redundancy in global connectivity. We formulate skeletal construction as a maximum-weight edge-disjoint spanning tree packing problem at the community level:
        \begin{equation}
        \begin{aligned}
        \mathop{\arg\max}\limits_{T_1,\dots,T_K}\quad & \sum_{t=1}^K \sum_{e\in E(T_t)} w_e \\
        \text{s.t.}\quad & T_t \text{ is a spanning tree of } \bar{G},\\
        & \forall t=1,\dots,K, \\
        & E(T_a) \cap E(T_b) = \emptyset,\ \forall a \neq b.
        \end{aligned}
        \label{eq:edst_packing}
        \end{equation}
        The union of the selected bridge edges $\bigcup_{t=1}^K E(T_t)$ provides redundant global routes across communities, acting as ``backup'' connections if some edges fail during incremental registration.
        
        We adopt a greedy Kruskal-style procedure~\cite{kruskal_shortest_1956} to approximate this objective while verifying candidate bridges using a strong two-view disambiguator (Doppelgangers++~\cite{xiangli_doppelgangers_2025}). Concretely, we traverse bridge candidates in descending weight order, and add a bridge only if (i) it connects two different disjoint-set-union (DSU) components at the community level, (ii) it satisfies a diversity constraint recorded by $\mathcal{U}$ in Alg.~\ref{alg:skeleton_build}, and (iii) it passes DG++ thresholding. This ensures that expensive verification is spent only on a small set of globally influential edges rather than on all pairs.
        
        \begin{algorithm}[t]
        \DontPrintSemicolon
        \caption{Skeletal Subgraph Building}
        \label{alg:skeleton_build}
        \KwIn{$G=(V,E,w)$; $\mathcal{P}$; bridges $\mathcal{B}$; $K$; threshold $\tau$}
        \KwOut{$G_s=(V_s,E_s)$}
        
        $E_s\leftarrow\emptyset$;\quad $\mathcal{U}\leftarrow\emptyset$ 
        
        \For{$t=1$ \KwTo $K$}{
          initialize DSU over communities $\mathcal{P}$;\quad $T_t\leftarrow\emptyset$\;
          \ForEach{$e_{ij}\in\mathcal{B}$ by descending $w_{ij} $}{
            let $\kappa$ be the community-pair induced by $e_{ij}$\;
            \lIf{$\kappa\notin\mathcal{U}$, $\mathcal{C}_a$ and $\mathcal{C}_b$ are in different DSU sets, and $\textsc{DG++}(i,j)\ge\tau$}{
              $T_t\leftarrow T_t\cup\{e_{ij}\}$;\ \textsc{Union}$(\mathcal{C}_a,\mathcal{C}_b)$
            }
            \lIf{$|T_t|=m-1$}{break}
          }
          $E_s\leftarrow E_s\cup T_t$;\quad $\mathcal{U}\leftarrow \mathcal{U}\cup\{\kappa(e)\mid e\in T_t\}$\;
        }
        $V_s\leftarrow\{v\in V \mid \exists\, e\in E_s \text{ incident to } v\}$\;
        \Return $G_s=(V_s,E_s)$\;
        \end{algorithm}

        \begin{figure}
            \centering
            \includegraphics[width=1\linewidth]{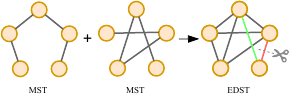}
            \caption{Packing multiple MST into EDST: \textit{2-edge-connectivity} guarantees graph to be connected for at most 1 edge failure, enhancing the robustness of the skeleton graph.}
            \label{fig:EDST}
        \end{figure}
        
        This construction yields a clean and streamlined skeleton spanning all communities while providing explicit redundancy. In particular, packing $K$ spanning trees at the community level implies controlled edge redundancy and supports robustness to bridge failures during incremental reconstruction (Fig.~\ref{fig:EDST}). Consider the community-level skeletal graph $\bar{G}_s = (\mathcal{P}, \bigcup_{t=1}^K E(T_t))$. Its edge-connectivity satisfies
        \begin{equation}
        \lambda(\bar{G}_s) := \min_{\emptyset \neq S \subsetneq \mathcal{P}} |\delta_{\bar{G}_s}(S)| \geq K,
        \end{equation}
        where $\delta_{\bar{G}_s}(S)$ denotes the cut-set between $S$ and $\mathcal{P}\setminus S$. Intuitively, each spanning tree must cross every non-trivial cut at least once, and edge-disjointness ensures at least $K$ distinct crossings. More generally, by the Tutte--Nash-Williams theorem~\cite{tutte_problem_1961,nash-williams_edge-disjoint_1961}, a graph admits $K$ edge-disjoint spanning trees if and only if every partition into $r\ge2$ parts has at least $K(r-1)$ crossing edges. Moreover, we optionally strengthen the skeleton with a small number of additional bridges, including top-weight bridges to densify connectivity, and verified backup bridges for low-weight connections. Since these added bridges are inter-community and potentially ambiguous, so we verify them. Finally, because an community may contribute multiple skeletal nodes incident to different bridges, we add intra-community edges to connect these skeletal nodes into a single connected subgraph within each community. These intra-community edges inherit the strong local support guaranteed by the top-$k$ construction and do not require verification.
    
    \subsection{Leaf Nodes Attachment}
        \label{leaf}
        Given the skeletal subgraph $G_s=(V_s,E_s)$, we next attach the remaining images in a way that preserves local support while avoiding unnecessary global densification. We define the \emph{leaf} node set as $V_\ell := V\setminus V_s$, so that $V = V_s \cup V_\ell$. For each community $\mathcal{C}_l\in\mathcal{P}$, we connect every leaf node $i\in \mathcal{C}_l\cap V_\ell$ to a small number of high-weight \emph{intra-community} neighbors. We prioritize edges from $i$ to already-selected skeletal nodes in the same community to anchor the leaf onto the scaffold, and optionally add a few additional leaf--leaf edges within the community for local support. Since communities are formed from strong local connectivity with large inlier counts, we do not apply DG++ to these intra-community attachment edges.
        
        The final hierarchical reconstruction graph is then defined as $H := (V, E_H)$ with $E_H := E_s \ \cup\ E_{\ell}$, where $E_{\ell}\subseteq \bigcup_{l=1}^m \big( (\mathcal{C}_l\cap V_\ell)\times \mathcal{C}_l \big)$ denotes the set of intra-community attachment edges incident to leaf nodes. In this hierarchy, $G_s$ serves as a compact, verified global scaffold across communities, while $E_{\ell}$ provides lightweight local edges that efficiently absorb remaining images without reintroducing costly global redundancy.
    
    \subsection{Coarse-to-Fine Reconstruction}
        \label{sec:recon}
        Given the hierarchical graph $H=(V,E_H)$ with skeleton $G_s=(V_s,E_s)$ and leaf attachments $E_{\ell}$, we perform a two-stage incremental SfM that follows the same hierarchy. We denote a sparse reconstruction by $\mathcal{R}=(\mathbf{P}, \mathbf{X})$, where $\mathbf{P}=\{\mathbf{P}_i\in\mathrm{SE}(3)\mid i\in V_{\mathcal{R}}\}$ is the set of estimated camera poses for the registered image set $V_{\mathcal{R}}\subseteq V$, and $\mathbf{X}$ is the set of triangulated 3D points.
        
        \textbf{Scaffold reconstruction.} We first run an off-the-shelf incremental SfM solver (COLMAP~\cite{schonberger_structure--motion_2016}) restricted to the skeletal subgraph $G_s$ to obtain a stable scaffold reconstruction $\mathcal{R}_s=(\mathbf{P}_s,\mathbf{X}_s)$. Because bridges in $E_s$ are selectively verified, the scaffold is designed to be robust to hard ambiguity and provides reliable global propagation across communities. Importantly, optimizing on this compact scaffold reduces the effective problem size early, alleviating the back-end cost that grows quickly with the number of cameras and constraints.
        
        \textbf{Efficient attachment.} Starting from the scaffold, we register the remaining images $V_\ell$ guided only by intra-community attachment edges. For an unregistered image $i\in V\setminus V_{\mathcal{R}}$, we collect its already-registered neighbors in $H$ via $E_{\ell}$, lift their 2D--2D matches to 2D--3D correspondences using existing points in $\mathbf{X}$, and estimate $\mathbf{P}_i$ with a robust PnP solver. Upon successful registration, we triangulate new points between $i$ and its registered neighbors and apply local bundle adjustment to refine the expanded model. We iterate this registration--triangulation loop until the number of registered cameras no longer increases, and finally run a global bundle adjustment over all poses and points to obtain the final reconstruction $\mathcal{R}=(\mathbf{P},\mathbf{X})$.
        
        This coarse-to-fine schedule establishes the globally critical, ambiguity-prone connections and is reconstructed first on a compact graph, while the remaining images are absorbed efficiently through reliable local support inside communities. As a result, HiSfM mitigates ambiguity-induced drift while reducing redundant computation in the back-end optimization.

\section{Experiments}
    \begin{table*}[]
    \Huge
    \centering
    \resizebox{\textwidth}{!}{%
    \begin{tabular}{@{}c|ccc|cccccc|cccccc|cccccc@{}}
    \toprule
     &
      \multicolumn{3}{c|}{COLMAP\cite{schonberger_structure--motion_2016}} &
      \multicolumn{6}{c|}{CamTrip\cite{manam_leveraging_2024}} &
      \multicolumn{6}{c|}{DG++\cite{xiangli_doppelgangers_2025}} &
      \multicolumn{6}{c}{Ours} \\ \cmidrule(l){2-22} 
    \multirow{-2}{*}{\textbf{Scene}} &
      \textsurd/× &
      \#reg.$\uparrow$ &
      time $\downarrow$ &
      \textsurd/× &
      \#reg.$\uparrow$ &
      \multicolumn{1}{l}{\#points $\uparrow$} &
      \multicolumn{1}{l}{err.$\downarrow$} &
      \multicolumn{1}{l}{len.$\uparrow$} &
      time $\downarrow$ &
      \textsurd/× &
      \#reg.$\uparrow$ &
      \multicolumn{1}{l}{\#points $\uparrow$} &
      \multicolumn{1}{l}{err.$\downarrow$} &
      \multicolumn{1}{l}{len.$\uparrow$} &
      time $\downarrow$ &
      \textsurd/× &
      \#reg.$\uparrow$ &
      \multicolumn{1}{l}{\#points $\uparrow$} &
      \multicolumn{1}{l}{err.$\downarrow$} &
      \multicolumn{1}{l}{len.$\uparrow$} &
      time $\downarrow$ \\ \midrule
    Alexander Nevsky Cathedral &
      {\color[HTML]{FF0000} ×} &
      447 &
      4.8 &
      {\color[HTML]{00B050} \textbf{\textsurd}} &
      425 &
      {\ul 99} &
      \textbf{0.8} &
      {\ul 7.8} &
      \textbf{3.3} &
      {\color[HTML]{00B050} \textbf{\textsurd}} &
      {\ul 447} &
      \textbf{100} &
      \textbf{0.8} &
      \textbf{7.9} &
      {\color[HTML]{FF0000} 130.2} &
      {\color[HTML]{00B050} \textbf{\textsurd}} &
      \textbf{448} &
      \textbf{100} &
      \textbf{0.8} &
      \textbf{7.9} &
      {\ul 3.6} \\
    Arc de Triomphe &
      {\color[HTML]{FF0000} ×} &
      425 &
      3.5 &
      {\color[HTML]{00B050} \textbf{\textsurd}} &
      371 &
      88 &
      \textbf{0.7} &
      \textbf{6.9} &
      \textbf{2.7} &
      {\color[HTML]{00B050} \textbf{\textsurd}} &
      {\ul 425} &
      \textbf{91} &
      \textbf{0.7} &
      {\ul 6.8} &
      65.7 &
      {\color[HTML]{00B050} \textbf{\textsurd}} &
      \textbf{428} &
      {\ul 89} &
      \textbf{0.7} &
      \textbf{6.9} &
      {\ul 8.3} \\
    Berliner Dom &
      {\color[HTML]{FF0000} ×} &
      1603 &
      180.3 &
      \multicolumn{6}{c|}{fail.} &
      {\color[HTML]{00B050} \textbf{\textsurd}} &
      \textbf{1606} &
      \textbf{258} &
      \textbf{0.7} &
      \textbf{16.0} &
      {\color[HTML]{FF0000} 598.8} &
      {\color[HTML]{00B050} \textbf{\textsurd}} &
      \textbf{1606} &
      {\ul 256} &
      {\ul 0.8} &
      \textbf{16.0} &
      \textbf{17.9} \\
    Big Ben &
      {\color[HTML]{FF0000} ×} &
      397 &
      3.3 &
      {\color[HTML]{00B050} \textbf{\textsurd}} &
      {\ul 365} &
      \textbf{82} &
      \textbf{0.7} &
      7.9 &
      \textbf{2.4} &
      {\color[HTML]{00B050} \textbf{\textsurd}} &
      \textbf{396} &
      \textbf{82} &
      \textbf{0.7} &
      {\ul 8.0} &
      {\color[HTML]{FF0000} 89.6} &
      {\color[HTML]{00B050} \textbf{\textsurd}} &
      394 &
      \textbf{82} &
      \textbf{0.7} &
      \textbf{8.1} &
      {\ul 3.1} \\
    Brandenburg Gate &
      {\color[HTML]{FF0000} ×} &
      173 &
      0.9 &
      {\color[HTML]{00B050} \textbf{\textsurd-}} &
      143 &
      {\ul 25} &
      0.9 &
      {\ul 8.7} &
      \textbf{0.8} &
      {\color[HTML]{00B050} \textbf{\textsurd}} &
      \uline{154}+\textbf{21} &
      \textbf{27} &
      \textbf{0.9} &
      8.4 &
      {\color[HTML]{FF0000} 39.5} &
      {\color[HTML]{00B050} \textbf{\textsurd-}} &
      \textbf{155} &
      {\ul 25} &
      \textbf{0.9} &
      \textbf{9.0} &
      {\ul 2.3} \\
    Church on Spilled Blood &
      {\color[HTML]{FF0000} ×} &
      274 &
      2.5 &
      {\color[HTML]{00B050} \textbf{\textsurd-}} &
      134+59 &
      {\ul 75} &
      \textbf{0.6} &
      7.9 &
      \textbf{1.3} &
      {\color[HTML]{00B050} \textbf{\textsurd}} &
      \textbf{157}+\textbf{106} &
      \textbf{76} &
      \textbf{0.6} &
      {\ul 8.2} &
      {\color[HTML]{FF0000} 69.8} &
      {\color[HTML]{00B050} \textbf{\textsurd}} &
      \uline{153}+\textbf{106} &
      73 &
      {\ul 0.7} &
      \textbf{8.4} &
      {\ul 4.3} \\
    Indoor &
      {\color[HTML]{FF0000} ×} &
      152 &
      0.7 &
      {\color[HTML]{00B050} \textbf{\textsurd}} &
      \textbf{152} &
      31 &
      \textbf{0.5} &
      {\ul 4.2} &
      \textbf{0.5} &
      {\color[HTML]{00B050} \textbf{\textsurd}} &
      \textbf{152} &
      \textbf{69} &
      {\ul 0.6} &
      \textbf{4.8} &
      {\color[HTML]{FF0000} 30.2} &
      {\color[HTML]{00B050} \textbf{\textsurd}} &
      \textbf{152} &
      \textbf{69} &
      {\ul 0.6} &
      \textbf{4.8} &
      {\ul 2.8} \\
    Radcliffe Camera &
      {\color[HTML]{FF0000} ×} &
      281 &
      2.6 &
      {\color[HTML]{00B050} \textbf{\textsurd-}} &
      172+46 &
      \textbf{81} &
      \textbf{0.7} &
      8.6 &
      \textbf{1.4} &
      {\color[HTML]{00B050} \textbf{\textsurd}} &
      \uline{186}+\textbf{94} &
      {\ul 80} &
      \textbf{0.7} &
      \textbf{8.8} &
      {\color[HTML]{FF0000} 64.1} &
      {\color[HTML]{00B050} \textbf{\textsurd}} &
      \textbf{188+94} &
      \textbf{81} &
      \textbf{0.7} &
      {\ul 8.7} &
      {\ul 8.4} \\ \midrule
    Books &
      {\color[HTML]{FF0000} ×} &
      21 &
      0.1 &
      {\color[HTML]{00B050} \textbf{\textsurd-}} &
      6+15 &
      \textbf{11} &
      \textbf{0.4} &
      {\ul 4.1} &
      0.3 &
      {\color[HTML]{00B050} \textbf{\textsurd}} &
      \textbf{21} &
      {\ul 9} &
      \textbf{0.4} &
      \textbf{5.3} &
      1.0 &
      {\color[HTML]{00B050} \textbf{\textsurd}} &
      \textbf{21} &
      {\ul 9} &
      \textbf{0.4} &
      \textbf{5.3} &
      \textbf{0.2} \\
    Cereal &
      {\color[HTML]{FF0000} ×} &
      25 &
      0.1 &
      {\color[HTML]{00B050} \textbf{\textsurd}} &
      14 &
      9 &
      \textbf{0.4} &
      3.7 &
      {\ul 0.3} &
      {\color[HTML]{FF0000} ×} &
      25 &
      {\ul 12} &
      \textbf{0.4} &
      \textbf{4.6} &
      1.4 &
      {\color[HTML]{00B050} \textbf{\textsurd}} &
      \textbf{25} &
      \textbf{13} &
      \textbf{0.4} &
      {\ul 4.4} &
      \textbf{0.2} \\
    Cup &
      {\color[HTML]{FF0000} ×} &
      64 &
      0.1 &
      {\color[HTML]{FF0000} ×} &
      64 &
      {\ul 7} &
      {\ul 0.7} &
      \textbf{7.3} &
      0.3 &
      {\color[HTML]{00B050} \textbf{\textsurd}} &
      \textbf{64} &
      \textbf{8} &
      0.6 &
      {\ul 6.2} &
      {\ul 8.8} &
      {\color[HTML]{00B050} \textbf{\textsurd}} &
      \textbf{64} &
      \textbf{8} &
      \textbf{0.4} &
      {\ul 6.1} &
      \textbf{0.4} \\
    Desk &
      {\color[HTML]{FF0000} ×} &
      31 &
      0.1 &
      {\color[HTML]{00B050} \textbf{\textsurd}} &
      \textbf{31} &
      \textbf{15} &
      \textbf{0.5} &
      {\ul 5.3} &
      \textbf{0.4} &
      {\color[HTML]{00B050} \textbf{\textsurd}} &
      \textbf{31} &
      \textbf{15} &
      \textbf{0.5} &
      \textbf{5.4} &
      1.3 &
      {\color[HTML]{00B050} \textbf{\textsurd}} &
      \textbf{31} &
      \textbf{15} &
      \textbf{0.5} &
      \textbf{5.4} &
      \textbf{0.2} \\
    Oats &
      {\color[HTML]{FF0000} ×} &
      23 &
      0.1 &
      {\color[HTML]{FF0000} ×} &
      23 &
      \textbf{9} &
      \textbf{0.4} &
      \textbf{5.7} &
      \textbf{0.3} &
      {\color[HTML]{FF0000} ×} &
      23 &
      \textbf{9} &
      \textbf{0.4} &
      \textbf{5.7} &
      {\ul 1.1} &
      {\color[HTML]{FF0000} ×} &
      23 &
      \textbf{9} &
      \textbf{0.3} &
      \textbf{5.7} &
      \textbf{0.3} \\
    Street &
      {\color[HTML]{FF0000} ×} &
      19 &
      0.1 &
      {\color[HTML]{00B050} \textbf{\textsurd}} &
      \textbf{19} &
      \textbf{5} &
      {\ul 0.5} &
      \textbf{3.1} &
      {\ul 0.3} &
      {\color[HTML]{00B050} \textbf{\textsurd}} &
      \textbf{19} &
      \textbf{5} &
      0.5 &
      \textbf{3.1} &
      0.7 &
      {\color[HTML]{00B050} \textbf{\textsurd}} &
      \textbf{19} &
      \textbf{5} &
      \textbf{0.4} &
      \textbf{3.1} &
      \textbf{0.1} \\
    Temple of Heaven &
      {\color[HTML]{FF0000} ×} &
      338 &
      6.7 &
      {\color[HTML]{00B050} \textbf{\textsurd}} &
      \textbf{338} &
      \textbf{192} &
      \textbf{0.9} &
      \textbf{9.2} &
      \textbf{5.2} &
      {\color[HTML]{00B050} \textbf{\textsurd}} &
      \textbf{338} &
      \textbf{192} &
      \textbf{0.9} &
      \textbf{9.2} &
      {\color[HTML]{FF0000} 80.3} &
      {\color[HTML]{00B050} \textbf{\textsurd}} &
      \textbf{338} &
      \textbf{192} &
      \textbf{0.9} &
      \textbf{9.2} &
      {\ul 6.8} \\ \bottomrule
    \end{tabular}%
    }
    \caption{{\color[HTML]{00B050}\textsurd}/\textcolor{red}{×} denotes correct/incorrect reconstruction. {\color[HTML]{00B050}\textsurd-} means oversplitting, and we ignore excessive reconstructed models with a number of registered images under 16 for Heinly \emph{et al.}~\cite{heinly_correcting_2014}. {\color[HTML]{00B050}\textsurd *} represents correct reconstruction yet with minor incorrection.``\#reg'.' is the number of reconstructed images. ``\#points'' is number of 3D points triangulated in $10^3$. ``err.'' is the mean reprojection error in pixels. ``len.'' is the mean track length of the reconstruction. ``time'' is the reconstruction runtime with inference included in minutes. \textbf{Bold} means the 1st place and \underline{underline} means 2nd place. \textcolor{red}{Red} in ``time'' column represents 10 times longer than other methods $\uparrow$:higher is better. $\downarrow$: lower is better.  Top rows are scenes from Heinly \emph{et al.}~\cite{heinly_correcting_2014} and bottom rows are from Yan \emph{et al.}~\cite{yan_distinguishing_2017}.}
    \label{tab:amb}
    \end{table*}

    We compare HiSfM against two representative lines of prior work discussed in Sec.~\ref{relate}: an efficiency-oriented view-graph method with ability of disambiguation, Camera Triplets (CamTrip)~\cite{manam_leveraging_2024}, and a robustness-oriented visual disambiguation method, Doppelgangers++ (DG++)~\cite{xiangli_doppelgangers_2025}. All methods are evaluated using the same SfM back-end COLMAP\cite{schonberger_structure--motion_2016} and the same feature extraction and matching outputs to ensure a fair comparison; differences arise only from how the view graph is selected/verified and how reconstruction is scheduled. 
    
    Experiments are conducted on a workstation with an Intel i7-14700K CPU, an NVIDIA RTX 3090 GPU (24GB), and 128GB RAM. We enable CUDA-accelerated bundle adjustment and use identical BA solver settings across methods. For CamTrip we use $m=0.7$, and for DG++ we use threshold $\tau=0.8$, following the authors' recommendations~\cite{manam_leveraging_2024,xiangli_doppelgangers_2025}. For HiSfM, we recommend setting the number of trees $K$ proportional to the image set size: using approximately 1 tree per 300 images to maintain sufficient global connectivity as the dataset grows. 
    In our experiments, $K$ is set to $\text{max}(1, \text{round}(\frac{n}{300}))$ where $n$ is the number of images in dataset.
    The ablation study of adopting such strategy is described in Sec. \ref{sec:ablation}. 
    Runtimes include the cost of disambiguation inference and reconstruction.

    \subsection{Evaluation on Ambiguous Datasets}

        \begin{figure*}
            \centering
            \includegraphics[width=0.95\linewidth]{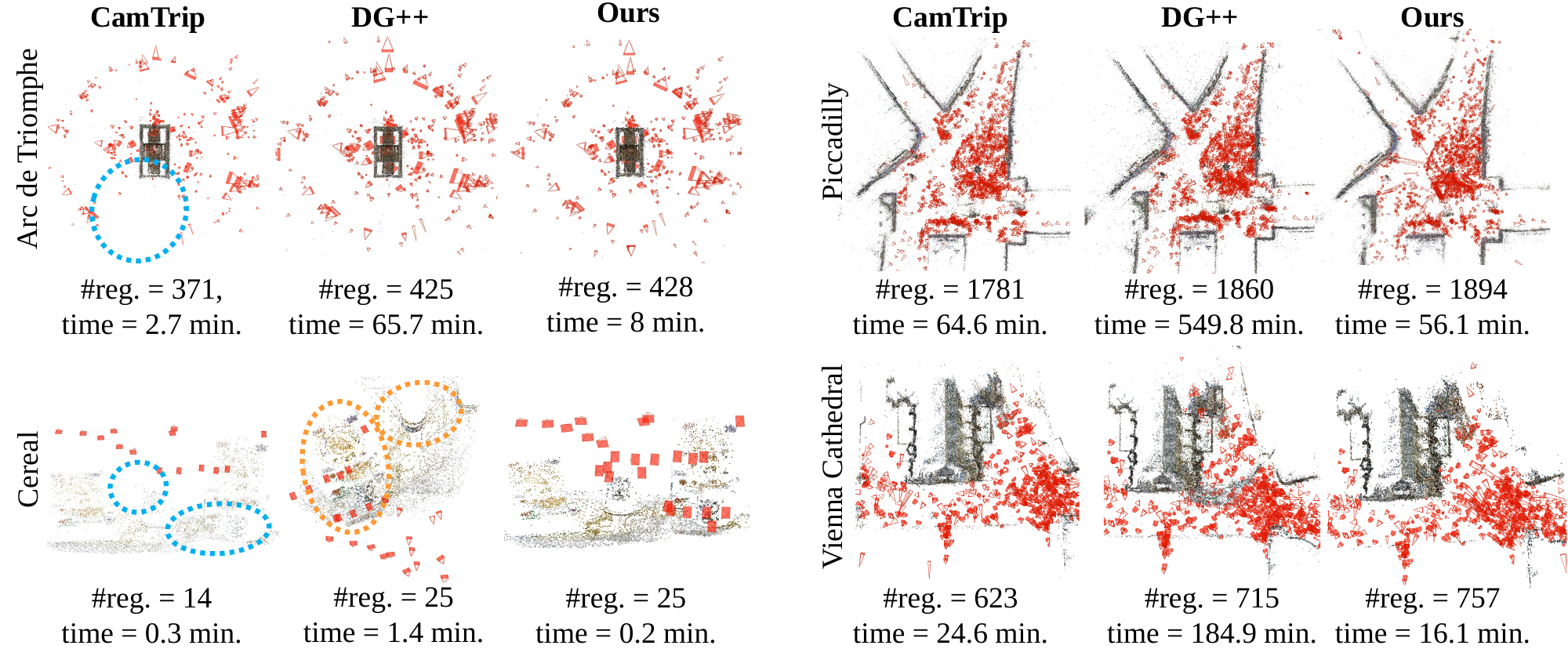}
            \caption{Comparison of the visualized results from Camtrip~\cite{manam_leveraging_2024}, DG++~\cite{xiangli_doppelgangers_2025} and our HiSfM for two ambiguous datasets and two general datasets. \textcolor{blue}{Blue circle}: incomplete reconstruction; \textcolor{orange}{Orange circle}: incorrect reconstruction.}
            \label{fig:amb_viz}
        \end{figure*}

        We first evaluate on widely used ambiguity stress tests from Heinly \emph{et al.}~\cite{heinly_correcting_2014} and Yan \emph{et al.}~\cite{yan_distinguishing_2017}, using vanilla COLMAP~\cite{schonberger_structure--motion_2016} as a baseline. 
        
        The quantitative reconstruction results are shown in Tab.~\ref{tab:amb}.     
        Overall, the results reflect the robustness--efficiency trade-off. The method CamTrip is typically fast for Heinly suite, but often registers fewer images and may oversplit the scene, consistent with aggressive sparsification that can discard globally important connections~\cite{manam_leveraging_2024}. 
        The method DG++ performs better than CamTrip, but incurs substantially higher runtime because it requires evaluating a large number of candidate pairs~\cite{xiangli_doppelgangers_2025}. In contrast, HiSfM achieves strong completeness while keeping runtime low: by verifying only inter-community bridges and reconstructing a compact scaffold before attaching remaining images, HiSfM prevents ambiguity-induced drift without paying the cost of exhaustive verification.
        
        A representative example is \textit{Cereal}, where DG++ fails to produce a correct reconstruction and CamTrip registers only about 56\% of images (Fig.~\ref{fig:amb_viz}), while HiSfM recovers a correct full registration and is also the fastest among compared methods. This case shows the advantage of our ``selective verification + representative skeleton'' design.
        
        To assess not only completeness but also geometric quality, we further evaluate the resulting camera poses by re-triangulating 3D points from the original (unfiltered) COLMAP database using the raw pairs and keypoint matches, and then measuring the total number of triangulated points, mean reprojection error, and mean track length. Since COLMAP filters triangulation outliers based on reprojection consistency, these statistics provide a practical proxy for pose quality under a shared triangulation pipeline. Across ambiguous scenes, HiSfM achieves comparable or better accuracy metrics than CamTrip and DG++ while substantially reducing runtime, supporting the effectiveness of our method.
        For example, on the \textit{Big Ben} dataset, our method registers a similar number of images as DG++, while being over 25 times faster. Although CamTrip achieves higher efficiency, it ignores too many edges in the view graph, leading to incomplete scene reconstruction.
        
    \subsection{Evaluation on General Datasets}
    
\begin{table}[]
\Huge
\centering
\resizebox{\columnwidth}{!}{%
\begin{tabular}{@{}c|c|cc|cc|cc@{}}
\toprule
 &
   &
  \multicolumn{2}{c|}{CamTrip\cite{manam_leveraging_2024}} &
  \multicolumn{2}{c|}{DG++\cite{xiangli_doppelgangers_2025}} &
  \multicolumn{2}{c}{Ours} \\ \cmidrule(l){3-8} 
\multirow{-2}{*}{\textbf{Scene}} &
  \multirow{-2}{*}{\#images} &
  \#reg.$\uparrow$ &
  time $\downarrow$ &
  \#reg.$\uparrow$ &
  time $\downarrow$ &
  \#reg.$\uparrow$ &
  time $\downarrow$ \\ \midrule
Alamo               & 627  & 477     & 11.9             & \textbf{573} & {\color[HTML]{FF0000} 177.6}  & {\ul 559}     & \textbf{10.3}  \\
Ellis community        & 247  & 200     & \textbf{1.2}     & \textbf{224} & {\color[HTML]{000000} 73.1}   & \textbf{224}  & {\ul 16.2}     \\
Gendarmenmarkt      & 742  & 543     & \textbf{5.6}     & {\ul 574}    & {\color[HTML]{000000} 127.3}  & \textbf{640}  & {\ul 16.0}     \\
Madrid Metropolis   & 394  & 69      & \textbf{1.2}     & {\ul 196}    & {\color[HTML]{FF0000} 100.3}  & \textbf{197}  & {\ul 10.2}     \\
Montreal Notre Dame & 474  & \multicolumn{2}{c|}{fail.} & {\ul 327}    & {\color[HTML]{FF0000} 122.4}  & \textbf{333}  & \textbf{9.3}   \\
NYC Library         & 376  & 252     & \textbf{1.9}     & {\ul 311}    & {\color[HTML]{FF0000} 61.5}   & \textbf{318}  & {\ul 5.9}      \\
Piazza del Popolo   & 354  & 259     & \textbf{2.0}     & {\ul 294}    & {\color[HTML]{FF0000} 82.7}   & \textbf{323}  & {\ul 4.2}      \\
Piccadilly          & 2508 & 1781    & {\ul 64.6}       & {\ul 1860}   & 549.8                         & \textbf{1894} & \textbf{56.1}  \\
Roman Forum         & 1084 & 844     & {\ul 13.4}       & {\ul 910}    & {\color[HTML]{FF0000} 153.0}  & \textbf{969}  & \textbf{13.0}  \\
Tower of London     & 508  & 310     & \textbf{2.9}     & {\ul 365}    & {\color[HTML]{000000} 67.5}   & \textbf{408}  & {\ul 8.5}      \\
Trafalgar           & 5058 & \multicolumn{2}{c|}{fail.} & {\ul 3485}   & {\color[HTML]{FF0000} 3035.3} & \textbf{4703} & \textbf{170.4} \\
Union Square        & 930  & 587     & \textbf{3.8}     & {\ul 665}    & {\color[HTML]{000000} 125.5}  & \textbf{675}  & {\ul 14.8}     \\
Vienna Cathedral    & 918  & 623     & {\ul 24.6}       & {\ul 715}    & {\color[HTML]{FF0000} 184.9}  & \textbf{757}  & \textbf{16.1}  \\
Yorkminster         & 458  & 319     & \textbf{5.4}     & \textbf{411} & {\color[HTML]{000000} 67.7}   & {\ul 406}     & {\ul 9.8}      \\ \midrule
Notre Dame          & 553  & 362     & \textbf{12.1}    & {\ul 474}    & {\color[HTML]{FF0000} 196.0}  & \textbf{492}  & {\ul 12.5}     \\ \bottomrule
\end{tabular}%
}
\caption{ ``\#img'' stands for the total number of images in the dataset. ``\#reg''. is the number of reconstructed images. ``time'' is reconstruction runtime with inference included in minutes.  \textbf{Bold} means the 1st place and \underline{underline} means 2nd place. \textcolor{red}{Red} in ``time'' column represents an order of magnitude longer than other methods. $\uparrow$:higher is better. $\downarrow$: lower is better. Top rows are scenes from 1DSfM~\cite{wilson_robust_2014} and the bottom row Notre Dame is from Photo Tourism~\cite{snavely_photo_2006}.}
\label{tab:general}
\end{table}

        Next, we evaluate the generalization of our method on larger, less curated Internet photo collections from 1DSfM~\cite{wilson_robust_2014} and Photo Tourism~\cite{snavely_photo_2006}.     
        The quantitative reconstruction results are shown in Tab. ~\ref{tab:general}.    

        On these general datasets, HiSfM consistently improves completeness over CamTrip on most scenes while remaining much faster than DG++. The method DG++ suffers from extremely high runtime on large image sets due to its reliance on broad pairwise disambiguation; in dense settings this cost can approach $\mathcal{O}(n^2)$ in the number of images when many candidate pairs are evaluated. In contrast, HiSfM limits the verification to a bounded number of globally critical bridges, which empirically grows roughly as $\mathcal{O}(Kn)$ with controlled constants, and thus scales more favorably with dataset size. 
        The method CamTrip often remains fast but can over-sparsify, leading to oversplitting and reduced registration; it also fails on very large scenes such as \textit{Trafalgar}, where traversing a massive number of triplets becomes impractical. These results shows that HiSfM strikes a better robustness--efficiency balance by combining a compact, verified scaffold with efficient coarse-to-fine completion, achieving scalable reconstruction without sacrificing quality or completeness. 
        
        Furthermore, results on both ambiguity-prone and general datasets show that HiSfM not only resolves visual ambiguities but also generalizes effectively to standard Internet photo collections. It maintains strong performance on large-scale scenes while preserving reconstruction completeness and structural integrity.

    \subsection{Ablation Studies}
        \label{sec:ablation}
        The parameter $K$, which controls the number of trees during the skeleton graph construction, is vital for our hierachical SfM.
        Hence in this section, we study the effect of the number of trees $K$ on reconstruction completeness. 
        
\begin{figure}
            \centering
            \includegraphics[width=1\linewidth]{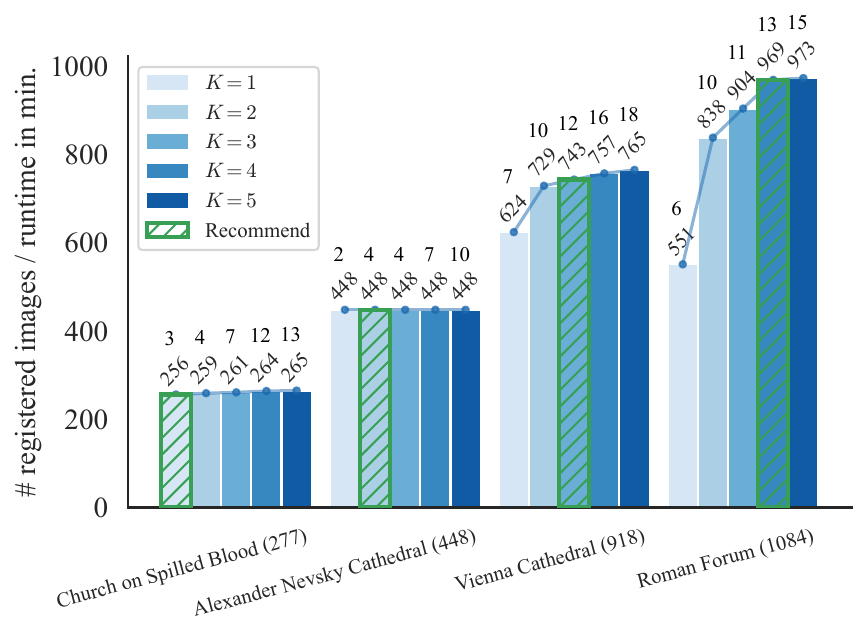}
            \caption{Ablation on the number of trees. Each bar group shows the number of registered images achieved with different $K$ on the same scene. Scene names and image set size are annotated as Scene(\#images). The registered image count is shown near the bar top, with the runtime in minutes annotated above it. Recommeded $K$ for each scene is denoted with green shaded bar.}
            \label{fig:ablation}
        \end{figure}

        We select representative scenes from ambiguous dataset Heinly \emph{et al.}~\cite{heinly_correcting_2014} and general dataset 1DSfM~\cite{wilson_robust_2014} with the image set size of roughly $\{300, 600, 900, 1200\}$ images, as multiples of 300, and run HiSfM with $K\in\{1,2,3,4, 5\}$. We measure completeness using the number of registered images. As shown in Fig.~\ref{fig:ablation}, increasing $K$ leads to a larger number of registered images, but also incurs higher runtime cost. Note that when $K\geq \text{max}(1, \text{round}(\frac{n}{300}))$, which is our recommended value for $K$, the number of registered images remains nearly stable. 
        For \textit{Roman Forum}, increasing $K$ from 4 to 5 introduces 15\% more runtime but yields only 4 additional images.
        This shows that our strategy of choosing the $K$ is sufficient to recover most images, while further increasing $K$ brings little improvement in terms of completeness but incurs additional runtime cost.
        
\section{Conclusion}
We presented HiSfM, a hierarchical coarse-to-fine structure-from-motion framework designed to address visual ambiguity and redundancy in real-world photo collections. HiSfM partitions the view graph into strong local communities, constructs a compact skeleton using edge-disjoint spanning trees, and applies verification only to critical inter-community bridges. By reconstructing a verified scaffold first and then attaching remaining images, HiSfM prevents ambiguity-induced drift while reducing redundant optimization. 
Experiments on ambiguous benchmarks and large Internet photo datasets show that HiSfM achieves robust and complete reconstructions with significantly lower runtime than exhaustive disambiguation methods and higher completeness than aggressive sparsification approaches. 

\section*{Acknowledgment}
This work was supported by the National Natural Science Foundation of China (Grant Numbers: U23A20386, 62572470, 62402494 and U22B2055).

\bibliographystyle{IEEEtran}
\bibliography{ref}

\end{document}